# ResSurv: New method proposed for survival prediction problem in cancer prognosis

Wankang Zhai
*EE-2, Electrical Engineering*
*International Collage*
*Dalian Maritime University*
zhai@dlmu.edu.cn

*Abstract*—Survival prediction is an important branch of cancer prognosis analysis. The model that predicts survival risk through TCGA genomics data can discover genes related to cancer and provide diagnosis and treatment recommendations based on patient characteristics. We found that deep learning models based on Cox proportional hazards often suffer from overfitting when dealing with high-throughput data. Moreover, we found that as the number of network layers increases, the experimental results will not get better, and network degradation will occur. Based on this problem, we propose a new framework based on Deep Residual Learning. Combine the ideas of Cox proportional hazards and Residual. And name it ResSurv. First, ResSurv is a feed-forward deep learning network stacked by multiple basic ResNet Blocks. In each ResNet Block, we add a Normalization Layer to prevent gradient disappearance and gradient explosion. Secondly, for the loss function of the neural network, we inherited the Cox proportional hazards methods, applied the semi-parametric of the CPH model to the neural network, combined with the partial likelihood model, established the loss function, and performed backpropagation and gradient update. Finally, we compared ResSurv networks of different depths and found that we can effectively extract high-dimensional features. Ablation experiments and comparative experiments prove that our model has reached SOTA(state of the art) in the field of deep learning, and our network can effectively extract deep information.

## I. Introduction

Predicting cancer through genetic analysis has been a recent focal point in research. Recent publications indicate that the integration of Artificial Neural Networks (ANN) has elevated the prognosis of cancer to a new dimension. However, given the distribution of TCGA data, relying on a limited number of samples may not sufficiently exploit high-dimensional features. We observed convergence challenges when applying Cox-nnet and Deepsurv for predicting TCGA mi-RNA dataset.

To address this issue, we propose a novel algorithm that combines meta-learning, data dimensionality reduction, and data augmentation methods to tackle the challenge of insufficient samples. Through experiments, our model successfully converged and surpassed other existing models in performance.

In the vast course of human history, cancer has always been a mysterious and formidable presence that evokes fear among people. With the rapid advancement of modern scientific technology, the survival rates of cancer patients have significantly improved. Cancer prognosis, as an indispensable component driving the development of the cancer field, is also flourishing. Cancer prognosis plays a crucial role, assisting physicians and patients in formulating more rational treatment plans, while also analyzing the causes of cancer occurrence.

Survival analysis stands out as the primary analytical method for cancer prognosis. It relies on statistical methods to analyze the survival time and status of patients, ultimately assessing the likelihood of survival after cancer diagnosis. Among these methods, the Cox model, also known as the proportional hazards model, is commonly employed. This model takes survival status (e) and survival time (t) as dependent variables and can simultaneously consider the impact of multiple covariates on the survival period.

With the advent of the era of deep learning, cancer prediction models based on deep learning have emerged in recent years. DeepSurv, a deep neural network, utilizes multiple hidden layers to extract features and combines them with Cox regression analysis. In comparison to Cox-nnet, DeepSurv enriches the number of hidden layers, providing the network with more parameters. Both models demonstrate satisfactory fitting and stable results on small datasets. However, when applied to high-dimensional TCGA gene datasets with features ranging from 10,000 to 20,000, a single network may struggle to fully fit the data, leading to overfitting, also known as the dimensionality catastrophe.

## II. Resources and Methods

In this chapter, We briefly introduce Cox propagation hazard, and made a necessary calculation about ResSurv.

### A. Dataset Preparation

In this work we are using 12 Datasets form TCGA database. These 15 datasets are KIRC: Kidney Renal Clear Cell Carcinoma LGG: Low-Grade Glioma LUAD: Lung Adenocarcinoma LUSC: Lung Squamous Cell Carcinoma PAAD: Pancreatic Adenocarcinoma SARC: Sarcoma STAD: Stomach Adenocarcinoma BRCA: Breast Invasive Carcinoma CESC: Cervical Squamous Cell Carcinoma and Endocervical Adenocarcinoma COAD: Colon Adenocarcinoma HNSC: Head and Neck Squamous Cell Carcinoma OV: Ovarian Serous

Cystadenocarcinoma LIHC: Liver Hepatocellular Carcinoma. For more details, see Appendix I.

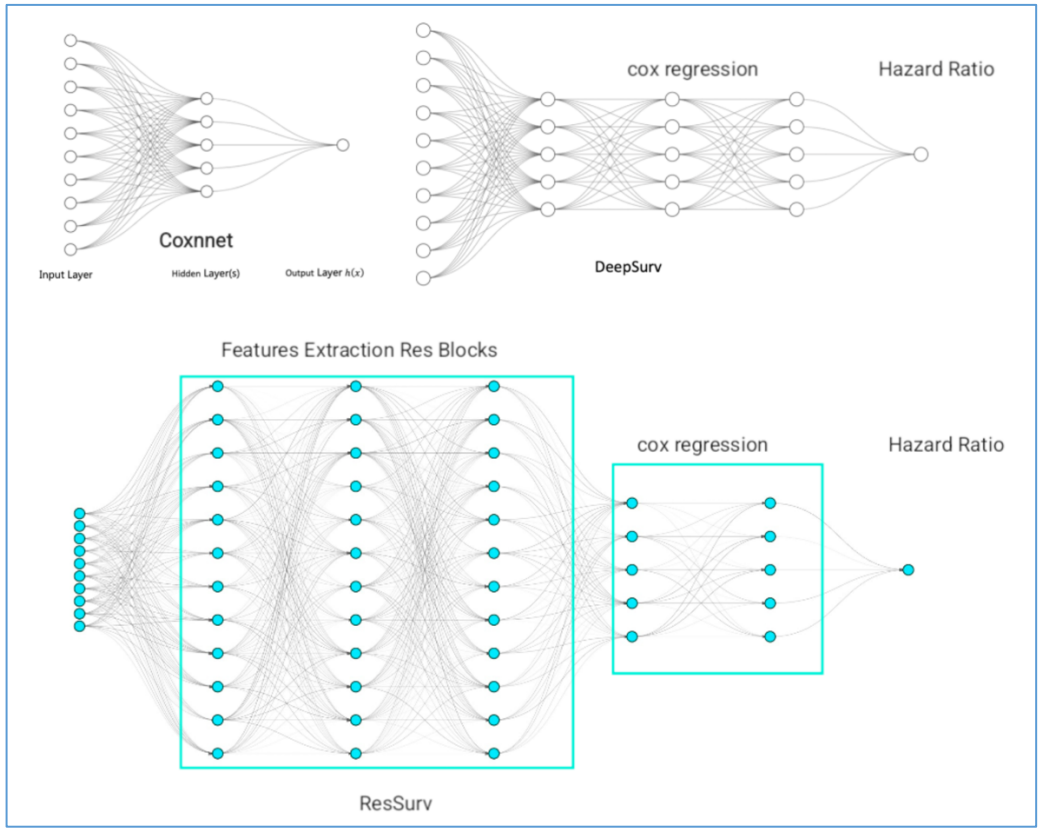

*Figrue 1-a the difference between Coxnnet and Deepsurv*

*Firgre 1-b Flowchart of ResSurv*

*B.Method*

*1) Cox hazard method*

In Survival analysis, we are interested in modeling the time versus a piece of patient information. One common approach to translation is to model survival time and patient information using semi-parametric risk models. The Cox model is a classic example based on the hazard function, commonly used in survival analysis. The hazard function is defined as:

$$\lambda(t) = \lim_{\delta \to 0} \frac{\mathbb{P}(t \leq T < t + \delta | T \geq t)}{\delta}$$

Where $\mathbb{P}$ is the probability of each individual surviving beyond t, and the hazard function indicates during a limited time period $\delta$, the survival rate of each person at a certain time. (DR Cox 1975) [1]

The hazard function can be rewrite as the following function:

$$\lambda(t|X) = \lambda_0(t) \cdot r(x)$$

$$r(x) = e^{h(x)} = e^{\beta^T x}$$

Where $h(x)$ represents the output value predicted by the model, and $\beta^T = (\beta_1, \ldots, \beta_p)$ is a vector of unknown parameters. And $\lambda_0(t)$ is an unknown arbitrary nonnegative function of time giving the hazard when $x = 0$.

Cox propagation hazards model (CPH) is the basic model for survival prediction. CPH is a semi–parametric model with estimating $r(x)$. We can find the probability of death to patients $i$ at time $T_i$ is:

$$L_n(\beta) = \frac{\lambda(T_i|X_i)}{\Sigma_{j:T_j \geq T_i}\lambda(T_i|X_j)} = \frac{\lambda_0(T_i)r(x_n)}{\Sigma_{j:T_j \geq T_i}\lambda_0(T_i)r(x_j)} = \frac{r(x_n)}{\Sigma_{j:T_j \geq T_i}r(x_j)}$$

Cox partial likelihood is the product of the probability at each event time $T_i$ that the event has occurred to the individual $i$. Consequently, Cox partial likelihood estimate $\beta$ without considering the time relevant parametric component $\lambda_0(t)$. Parameters is estimated by log-partial likelihood.

$$\mathcal{L}(\beta) = -\frac{1}{N_{E=1}} \sum_{i:E_i=1} \left( \beta^T x_i - \log \sum_{j=1}^{n} Y_j(T_i) e^{\beta^T x_j} \right)$$

Where $Y_j(t) = 1$ $(T_i > t)$. This Loss function gives a fundamental method for Deep Learning. All the Cox Partial

likelihood models are based on this function. Our model ResSurv is also based on this loss function.

*2) ResNet Model*

Deep Residual Network proposes an optimization of learning residual functions, using nested function classes through identity mapping to ensure that each gradient update can move closer to the global optimum.

Residual Networks (ResNet) is proposed by (Kai et al.2016)[ResNet]. It was originally proposed to address two types of problems. One of them is the gradient vanishing and gradient exploding problem. Another one is Models degrade. We find these problems are common for high-throughput problems. High throughput data is likely to cause dimensional disaster. In high-dimensional space, the number of samples is small relative to the number of features. This makes the model susceptible to noise which makes it difficult for the model to generalize to unseen data.

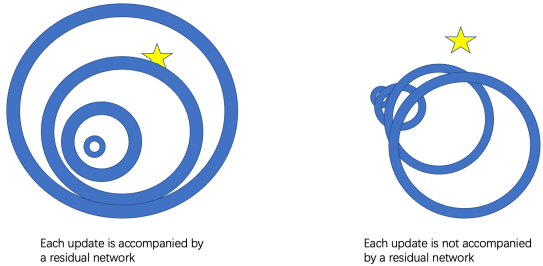

Figure 2 Different nested function classes

As seen in Figure 2, therefore, we can be sure to improve the performance of more complex function classes only if they contain smaller function classes. For deep neural networks, if we can train the newly added layers into identity functions, the new model will be equally effective as the original model. At the same time, adding layers seems to make it easier to reduce training error since the new model may lead to better solutions to fit the training data set. It won the 2015 ImageNet Image Recognition Challenge and profoundly influenced the design of subsequent deep neural networks. The core idea of residual networks is that each additional layer should more easily contain the original function as one of its elements. Thus, residual blocks (residual blocks) were born. This design had a profound impact on how to build deep neural networks.

*3) ResSurv Model*

ResSurv is a network for survival prediction. Combined with two parts. The main channel is calculated by a few stacked layers. The shortcut channel is calculated directly by a fully connected layer. For the whole model, we designed the layers as shown in Figure 1.

Identify applicable funding agency here. If none, delete this text box.

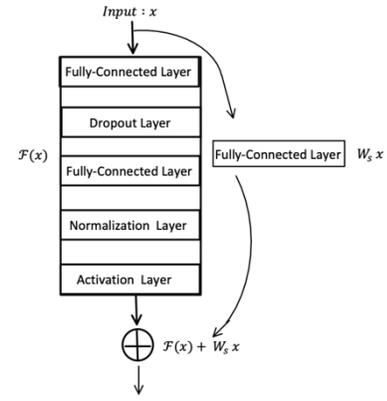

Figure 3 the structure of ResSurv Block

The output of each Block is defined as:

$$y = \mathcal{F}(x, \{W_i\}) + W_s x$$

Where the $x$ and $y$ are the input and output respectively. We can find the output combined with the Linear layers and nonlinear layers. Now we want to learn the residual $y - W_s x$, and the original mapping is recast into $\mathcal{F}(x, \{W_i\}) + W_s x$. The reason is that in high throughput data, the stack layers are hard to avoid gradient disappears.

$$\frac{\partial l}{\partial x} = \frac{\partial y}{\partial x}\frac{\partial l}{\partial y} = \left(\frac{\partial \mathcal{F}(x,\{W_i\})+W_s x}{\partial x}\right)\frac{\partial l}{\partial y} = \left(\frac{\partial \mathcal{F}(x,\{W_i\})}{\partial x} + W_s\right)\frac{\partial l}{\partial y}$$

Where the loss gradient contains two parts. When $\frac{\partial \mathcal{F}(x,\{W_i\})}{\partial x}$ is almost zero, the loss gradient can be approximated as $W_s \frac{\partial l}{\partial y}$. Back to the loss function, after adding a residual network, it can prevent this term from being the same.

$$log \sum_{j=1}^{n} Y_j(T_i) e^{\beta^T x_j}$$

In ResSurv, we are using some skills to overcome the overfitting problem. They are $\ell_2$ regression.

$$\mathcal{L}(\beta) = -\frac{1}{N_{E=1}}\sum_{i:E_i=1}\left(\beta^T x_i - log \sum_{j=1}^{n} Y_j(T_i) e^{\beta^T x_j}\right) + \lambda\|\beta\|_2^2$$

What's more, we are adding a Batch Normalization layer between blocks. Because it can reduce the dependence on parameter initialization, and increase the model generalization ability.

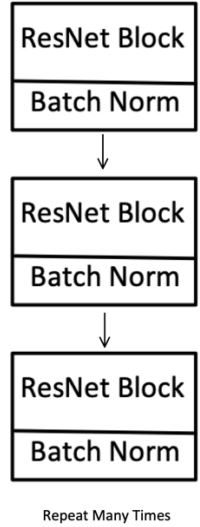

Figure4 the structure of ResSurv Network

In this work, we consider the batch normalization for ResSurv, shown in Figure4. Both input and output are two-dimensional tensors, we respect to $I_{f,n}$ and $O_{f,n}$ respectively. $f$ represents the number of features and $n$ represents the number of samples. Given the function, we can calculate the normalized layer:

$$O_{f,n} = \gamma \cdot \frac{I_{f,n} - \mu_n}{\sqrt{\sigma_n^2 + \epsilon}} + \beta \qquad \forall f, n$$

In this function, where $\mu_n$ and $\sigma_n^2$ are the mean and variance in $n$ dimension.

## III. EVALUATION METRICS

In this chapter, we compare the differences between ResSurv and other deep learning models in detail and conclude that our model performs better on 10 of the 12 datasets.

### A. Experiments Settings

We implemented our model using the PyTorch framework. We train and validate our model using an Nvidia 4090 (16GB) GPU. Firstly, we are using Grid Searching to find the best performance of the combination of parameters. And then with the best hyper parameter, we use inner 5-fold cross-validation to find the performance of our model. All the network was done by the same settings. We are adding early stopping for avoiding overfitting, and adjust the block layers from 5-7, adjusting learning rate and decaying rate, etc. we use the same dataset settings to train and evaluate Cox-nnet, DeepSurv, and XGBENC.

### B. Evaluation matric

In this work, we are mainly use the concordance index, which It is a statistical metric used to evaluate the predictive accuracy of survival analysis models. In survival analysis, C-index is usually used to evaluate the model's ability to rank survival times, that is, the accuracy of the model in predicting the probability of an event (such as death).

### C. Results

In Figure 5, we show the direct figure comparation between 4 models. You can derive the Hyperparameters and c-index numeric values from Appendix II .

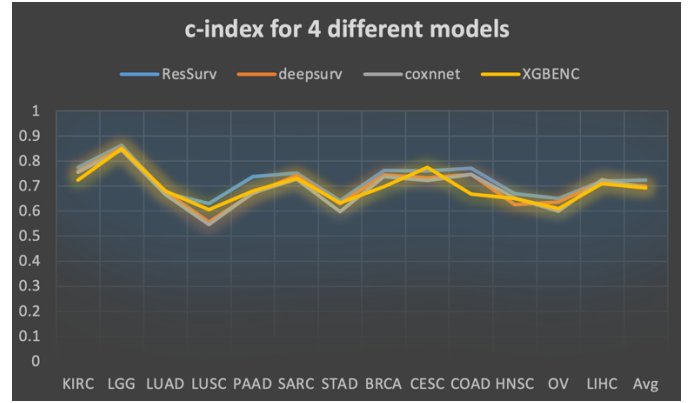

Figure 5 C-index for 4 differents models

## IV. DISCUSSION

Cancer survival prediction stands as a pivotal segment within cancer prognosis analysis, furnishing insights crucial for diagnosis, treatment decisions, and treatment evaluation. Leveraging genomic data from reputable sources like TCGA, predictive models emerge as potent tools for pinpointing cancer-associated genes, profoundly impacting basic medical research. However, conventional deep learning models, particularly those anchored in Cox proportional hazards, often grapple with overfitting quandaries when confronted with extensive high-throughput datasets. Notably, we observed a tendency for predicted risk scores to stabilize with escalating epochs. Moreover, augmenting the number of network layers frequently fails to yield commensurate enhancements, potentially undermining network efficacy.

In tackling these hurdles, the ResSurv framework draws inspiration from the efficacy of deep residual learning, offering a promising remedy. Pioneering a novel approach, ResSurv amalgamates Cox proportional hazards with the residual principle, mitigating overfitting to a certain degree and enhancing model robustness.

### A. Loss function

The loss function in ResSurv is inherited from the Cox likelihood risk estimation method, seamlessly integrating the semi-parametric aspects of the CPH model into the neural network paradigm. By incorporating a partial likelihood model, ResSurv builds a loss function that includes a residual term. This enables efficient backpropagation and gradient updates, promoting model convergence and optimization.

*B. Architecture*

ResSurv adopts a feed-forward deep learning architecture comprising multiple basic ResNet Blocks. These blocks are enriched with Normalization Layers to counteract issues like gradient disappearance and explosion. This strategic integration fosters stability and enhances the model's ability to extract relevant features from complex data.

*C. Performance Evaluation*

Empirical evaluation of ResSurv unveils its efficacy in extracting high-dimensional features. Through ablation experiments and comparative analyses, ResSurv demonstrates superiority, attaining state-of-the-art status in the realm of deep learning for cancer prognosis analysis. Notably, the model's adeptness in harnessing deep information underscores its potential in clinical settings.

data, providing a promising avenue for advancing cancer prognostic analysis and personalized medicine.

## VI. AVAILABILITY OF DATA AND MATERIALS

The results shown here are in whole or part based upon data generated by the TCGA Research Network: https://www.cancer.gov/tcga.

## VII. AVAILABILITY OF CODES

Code is available at : https://github.com/Madrigalpp/Torch-version-for-TCGA-data-DeepSurv-

## V. CONCLUSION

Survival prediction is a key aspect of cancer prognostic analysis and provides insights into the impact of genes on cancer based on TCGA genomic data. By leveraging predictive models, we can discover genes associated with cancer and tailor diagnostic and treatment strategies to individual patients. However, traditional deep learning models based on Cox proportional hazards often encounter challenges such as overfitting, especially when processing high-throughput data. In addition, increasing the number of network layers does not necessarily improve performance, but may lead to a decrease in network performance.

To address these issues, we propose ResSurv, a novel framework that incorporates ideas from Cox proportional hazards and deep residual learning. ResSurv consists of a feed-forward deep learning network built from multiple basic ResNet blocks, each of which contains a normalization layer to alleviate gradient issues. We integrate the semiparametric nature of the Cox proportional hazards model into the loss function of the neural network and combine it with the partial likelihood model to achieve efficient backpropagation and gradient updating.

Through extensive experiments, including comparisons of ResSurv networks of different depths, we demonstrate the model's ability to effectively extract high-dimensional features. Ablation and comparative studies validate our approach, demonstrating ResSurv's state-of-the-art performance achievements in deep learning. Our framework is a powerful solution for extracting deep information from complex genomic

APPENDIX I

Here are the Data for my experiments

| Cohort | Disease name | Number of mRNA features provided | Number of patients alive | Number of patients dead | Number of patients with censored survival data after filtering | Number of patients with uncensored survival data after filtering | Number of mRNA features provided after filtering |
|---|---|---|---|---|---|---|---|
| BRCA | Breast Invasive Carcin | 19754 | 867 | 140 | 16490 | 864 | 134 |
| CESC | Cervival Squamous Ce | 19754 | 236 | 71 | 16246 | 233 | 71 |
| COAD | Colon Adenocarcinon | 19754 | 366 | 98 | 15937 | 361 | 97 |
| HNSC | Head and Neck Squar | 19754 | 304 | 223 | 16321 | 281 | 216 |
| KIRC | Kidney Renal Clear Ce | 19754 | 371 | 176 | 16594 | 366 | 172 |
| LGG | Brain Lower Grade Gl | 19754 | 388 | 125 | 16772 | 383 | 125 |
| LIHC | Liver Hepatocellular C | 19754 | 244 | 132 | 15580 | 230 | 128 |
| LUAD | Lung Adenocarcinoma | 19754 | 349 | 196 | 16537 | 188 | 336 |
| LUSC | Lung Squamous Cell ( | 19754 | 299 | 213 | 16759 | 286 | 210 |
| OV | Ovarian Serous Cystac | 19754 | 234 | 349 | 16765 | 143 | 230 |
| PAAD | Panceatic Adenocarci | 19754 | 85 | 100 | 16861 | 85 | 92 |
| STAD | Stomach Adenocarcin | 19754 | 270 | 168 | 17023 | 227 | 144 |

APPENDIX II

Here are Hyper Parameters and numeric values for my experiments.

| Hyper-Parameters | | | | | | |
|---|---|---|---|---|---|---|
| Optimizer | Adam | AdamW | SGD | | | |
| Activation | tanh | SELU | ReLU | | | |
| Dense Layers | 3 | 4 | 5 | 6 | 7 | |
| Nodes | 64 | 128 | 512 | 1024 | | |
| Learning Rate | 1.00E-01 | 1.00E-02 | 1.00E-03 | 1.00E-04 | | |
| L2 Regression | 2 | 4 | 6 | 8 | 10 | |
| Dropout | 0.2 | 0.4 | 0.6 | | | |
| LR Decay | 1.00E-02 | 1.00E-03 | 1.00E-04 | 1.00E-05 | | |

| c-index | ResSurv | deepsurv | coxnnet | XGBENC |
|---|---|---|---|---|
| KIRC | **0.7745** | 0.7586 | 0.75483 | 0.7237 |
| LGG | **0.8632** | 0.853317 | 0.843412 | 0.8485 |
| LUAD | **0.6742** | 0.676489 | 0.665532 | 0.6805 |
| LUSC | **0.6301** | 0.556426 | 0.547116 | 0.6058 |
| PAAD | **0.7375** | 0.67064 | 0.672963 | 0.6799 |
| SARC | **0.7514** | 0.742971 | 0.72789 | 0.7306 |
| STAD | **0.6423** | 0.628756 | 0.597826 | 0.6333 |
| BRCA | **0.7622** | 0.744763 | 0.738305 | 0.6983 |
| CESC | 0.7601 | 0.732233 | 0.72257 | **0.7744** |
| COAD | **0.7714** | 0.747728 | 0.747542 | 0.668 |
| HNSC | **0.6702** | 0.626026 | 0.653107 | 0.6504 |
| OV | **0.6507** | 0.635878 | 0.600393 | 0.6098 |
| LIHC | 0.7212 | 0.717844 | **0.726313** | 0.7109 |
| Avg | 0.723769231 | 0.69935931 | 0.692138385 | 0.69339231 |